\documentclass[10pt,twocolumn,letterpaper]{article}

\usepackage[pagenumbers]{iccv} 

\definecolor{iccvblue}{rgb}{0.21,0.49,0.74}
\usepackage[pagebackref,breaklinks,colorlinks,allcolors=iccvblue]{hyperref}

\usepackage{multirow}
\usepackage{multicol}
\usepackage{amssymb}
\usepackage{colortbl}
\usepackage{bm} 
\usepackage{pifont}

\definecolor{darkgreen}{RGB}{34,139,34}

\newcommand{\cmark}{\ding{51}}%
\newcommand{\xmark}{\ding{55}}%

\def\paperID{10104} 
\def\confName{ICCV}
\def\confYear{2025}

\title{CutS3D: Cutting Semantics in 3D for 2D Unsupervised Instance Segmentation}

\author{Leon Sick$^1$ \and Dominik Engel$^{1,2}$ \and Sebastian Hartwig$^1$ \and Pedro Hermosilla$^3$ \and Timo Ropinski$^1$\\
\\
$^1$Ulm University \quad $^2$KAUST \quad $^3$TU Vienna\\
}

\makeatletter
\def\and{\hskip 0.5em\@plus.17fil\relax}
\makeatother

\begin{document}
 \maketitle
 \begin{abstract}
Traditionally, algorithms that learn to segment object instances in 2D images have heavily relied on large amounts of human-annotated  data. Only recently, novel approaches have emerged tackling this problem in an unsupervised fashion. Generally, these approaches first generate pseudo-masks and then train a class-agnostic detector. While such methods deliver the current state of the art, they often fail to correctly separate instances overlapping in 2D image space since only semantics are considered. To tackle this issue, we instead propose to cut the semantic masks in 3D to obtain the final 2D instances by utilizing a point cloud representation of the scene. Furthermore, we derive a Spatial Importance function, which we use to resharpen the semantics along the 3D borders of instances. Nevertheless, these pseudo-masks are still subject to mask ambiguity. To address this issue, we further propose to augment the training of a class-agnostic detector with three Spatial Confidence components aiming to isolate a clean learning signal. With these contributions, our approach outperforms competing methods across multiple standard benchmarks for unsupervised instance segmentation and object detection. \\
Project Page: \href{https://leonsick.github.io/cuts3d/}{leonsick.github.io/cuts3d}
\vspace{-1mm}

\end{abstract}    
 \vspace{-4mm}
\section{Introduction}
\label{sec:intro}
Segmenting instances in natural images has long been considered a challenging task in computer vision. In recent years however, supervised algorithms~\cite{he2017mask, cai2019cascade, chen2019hybrid, hu2018learning, zhang2020mask, shen2021dct, cheng2022masked} have made impressive progress towards enabling the segmentation of a large diversity of objects in a variety of environments and domains. But for these to perform well, they require a large amount of labeled training data, which, especially for instance segmentation, is costly to obtain, since annotations have to contain information about individual instances of a particular class. 
\begin{figure}[!thb]
  \centering
   \includegraphics[width=\linewidth, page=4]{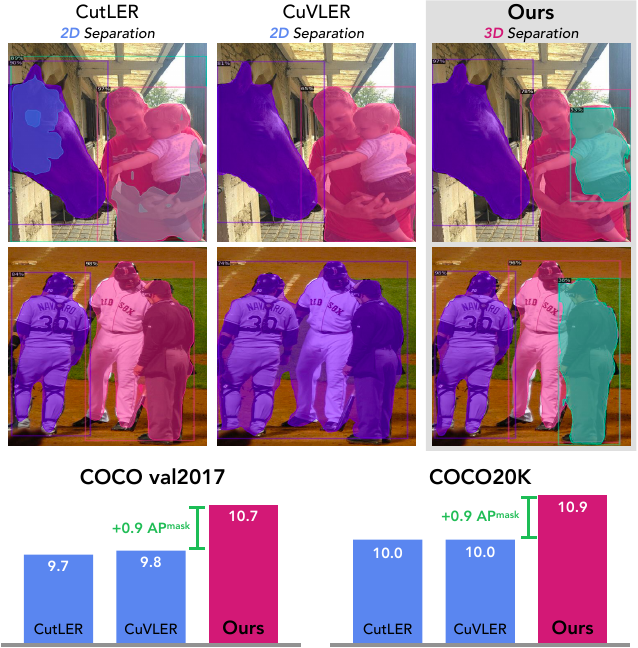}
   \vspace{-6mm}
   \caption{\textbf{Cutting Semantics Into Instances in 3D.} We leverage 3D information to separate semantics into instances to generate pseudo-masks on IN1K, then train a class-agnostic detector on them. The resulting model is able to separate instances with improved accuracy, outperforming previous approaches.  
}
   \label{fig:teaser}
   \vspace{-4.5mm}
\end{figure}
One prominent example is the COCO dataset~\cite{lin2014coco}, containing around 164 thousand images with instance mask annotations distributed across 80 classes, which required more than 28 thousand hours of annotation time across a large group of labelers. 
This enormous effort has sparked the field of unsupervised instance segmentation, which aims to develop algorithms that can perform such segmentations with similar quality, but without needing any human annotations used during training.
CutLER, a recent approach by Wang et al.~\cite{wang2023cut}, has made significant progress in unsupervised instance segmentation of natural scenes by training a class-agnostic detection network (CAD) on pseudo-masks generated by MaskCut. To do so, the authors leverage self-supervised features to capture the 2D semantics of the scene in an affinity graph and extract instance pseudo-masks. However, since their approach only considers semantic relations, it fails to separate instances of the same class that are overlapping or connected in 2D space. 
In contrast, humans inherently perceive the natural world in 3D, an ability which helps them to better separate instances by not just considering visual similarity, but also their boundaries in 3D. Nowadays, precise 3D information is readily available through zero-shot monocular depth estimators trained solely with sensor data and no human annotations, thus not breaking the unsupervised setting. 
Further, previous work has already shown that using spatial information is helpful for unsupervised semantic segmentation~\cite{sick2024unsupervised}. We argue that, in order to separate instances effectively, information about the 3D geometry of the scene also needs to be taken into account.

Within this work, we propose \textbf{CutS3D} (\textbf{Cut}ting \textbf{S}emantics in \textbf{3D}), the first approach to introduce 3D information to separate instances from semantics for 2D unsupervised instance segmentation. \textbf{CutS3D} effectively incorporates this 3D information in various stages:
For pseudo-mask extraction, we go beyond 2D separation and cut instances in 3D. Starting from an initial semantics-based segmentation, we cut the object along its actual 3D boundaries from a point cloud of the scene, obtained by orthographically unprojecting the depth map. To 3D-align the feature-based affinity graph for the initial semantic cut, we further compute a Spatial Importance function which assigns a high importance to regions with high-frequency depth changes. 
The Spatial Importance map is then used to sharpen the semantic affinity graph, effectively enriching it with 3D information in order to make cuts along object boundaries more likely.
While the generated pseudo ground truth is useful to train the CAD, these pseudo mask are inherently ambiguous. We therefore again leverage 3D information to extract information about the quality of the pseudo-masks at individual patches to separate clean from potentially noisy learning signals.
To achieve this, we introduce Spatial Confidence maps which are computed by performing multiple 3D cuts at different scales. We argue this captures the confidence the algorithm has in the instance separation along the spatial borders, since we observe that only for objects with unclear borders, these cuts yields slightly different results at varying scales.
We leverage these Spatial Confidence maps in three ways when training the CAD. First, we select only the highest-confidence masks within an image for copy-paste augmentation. Second, we perform alpha-blending so that object regions are blended with the current image proportionally to their region confidence.
Third, we use our Spatial Confidence maps to introduce a Spatial Confidence Soft Target loss, a more precise way to incorporate the signal quality in the mask loss. 
In summary, we propose the following contributions:
\begin{enumerate}[label=\arabic*.]
    \item \textbf{LocalCut} to cut objects along their actual 3D boundaries based on an initial semantic mask.

    \item \textbf{Spatial Importance Sharpening} for 3D-infusing the semantic graph to better represent 3D object boundaries.

    \item \textbf{Spatial Confidence} that captures pseudo-mask quality to select only \textbf{confident masks} for copy-paste augmentation, \textbf{alpha-blend} instances when copy-pasting, and introduce a novel \textbf{Spatial Confidence Soft Target Loss}.

\end{enumerate}

\begin{figure*}[!ht]
  \centering
    \includegraphics[width=\textwidth, page=1]{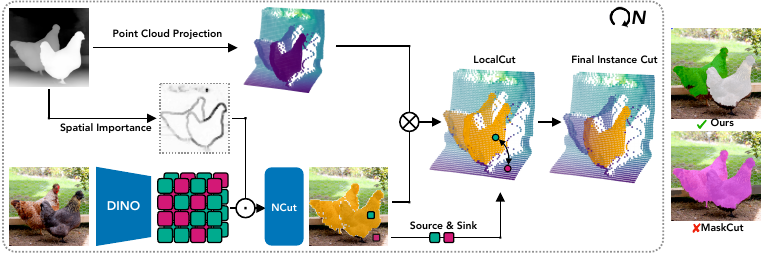}
    \caption{\textbf{CutS3D Pseudo-Mask Extraction Pipeline.} We separate instances in 3D, cutting semantics groups into instances even if they are connected in 2D space. To make the semantic affinity matrix 3D-aware, we sharpen it using Spatial Importance maps to improve the semantic relations along the 3D boundaries of instances.} 
    \label{fig:pseudo-mask}
    \vspace{-4mm}
\end{figure*}

 \section{Related Work}
\label{sec:related-work}

\noindent\textbf{Self-Supervised Representation Learning.} The advances of self-supervised feature learning, which aims to propose methods for learning deep features without human annotations or supervision, have been central to progress in unsupervised instance segmentation. One popular approach, DINO by Caron et al.~\cite{caron2021emerging}, employs a student-teacher learning process with cropped images to make a ViT learn useful features. Their work has been demonstrated to produce semantic features and attention maps, and has been employed by various approaches in unsupervised segmentation~\cite{simeoni2021lost, wang2023cut, arica2024cuvler, li2024promerge, sick2024unsupervised, hamilton2022unsupervised, seong2023leveraging, kim2024eagle}. In their work on MoCo, He et al.~\cite{he2020momentum} use contrastive learning and use a momentum encoder to achieve a more scalable and efficient pre-training process. DenseCL, proposed by Wang et al.~\cite{wang2021dense}, also utilizes contrastive learning and introduces a pairwise pixel-level similarity loss between image views.
Oquab et al.~\cite{oquab2023dinov2} propose DINO's successor, called DINOv2, by introducing optimizations such as KoLeo regulization~\cite{sablayrolles2019spreading}, Sinkhorn-Knopp centering~\cite{caron2020swav} or the patch-level objective from iBOT~\cite{zhouibot}. Another recent follow-up to DINO is proposed by Lui et al.~\cite{liu2024diffncuts} with DiffNCuts. They implement a differentiable version of Normalized Cuts (NCut) and use a mask consistency objective on cuts from different image crops to finetune a DINO model. Their model improves upon vanilla DINO for object discovery tasks.

\noindent\textbf{Unsupervised Instance Segmentation.} Recent progress in unsupervised instance segmentation has been accelerating rapidly.  MaskDistill~\cite{van2022maskdistill} uses a bottom-up approach and leverages a pixel grouping prior from MoCo features to extract masks, which serve as input to training a Mask R-CNN~\cite{he2017mask}. In contrast, FreeSOLO~\cite{wang2022freesolo} uses features from DenseCL as part of a key-query design to extract query-informed attention maps that represent masks, which are used to train a SOLO model~\cite{wang2021solo}. Wang et al.~\cite{wang2023cut} propose CutLER by extending an NCut-based mask extraction process from DINO features, first proposed by TokenCut~\cite{wang2023tokencut}, to the multi-instance case. After extracting pseudo-masks with MaskCut, they train a detector on them and for multiple rounds on its own predictions. We provide a more detailed description of CutLER in Section~\ref{subsec:preliminaries}. Wang et al. leverage CutLER with a divide-and-conquer approach to train UnSAM~\cite{wang2025unsam}. This unsupervised version of the segment anything model (SAM) \cite{kirillov2023segment} is tasked with detecting as many objects as possible with different granularity, as opposed to segmenting instances precisely like other related approaches.
CuVLER by Arica et al.~\cite{arica2024cuvler} use an ensemble of 6 DINO models to extract pseudo-masks, as well as a soft target loss for training of the detector. Finally, Li et al. propose ProMerge~\cite{li2024promerge} and attempt to identify the background of the scene in combination with point prompting to then sequentially segment and merge individual instances, then train a detection network on the instance masks.
 \section{Method}
\label{sec:method}
We build upon CutLER~\cite{wang2023cut} and first extract pseudo-masks to then train a CAD on this pseudo ground-thruth. We introduce LocalCut (Sec.~\ref{sec:localcut}) and Spatial Importance Sharpening (Sec.~\ref{sec:spatial-importance}) for pseudo-mask extraction, plus three Spatial Confidence components (Sec.~\ref{sec:spatial-confidence}) for training the CAD. 

\subsection{Preliminaries}
\label{subsec:preliminaries}
Following Wang et al.~\cite{wang2023cut}, we first feed the input image through a DINO~\cite{caron2021emerging} Vision Transformer (ViT)~\cite{dosovitskiy2020vit} $\mathcal{F}: \mathbb{R}^{3\times H_\text{in}\times W_\text{in}} \rightarrow \mathbb{R}^{C\times H\times W}$ to extract a feature map $f \in \mathbb{R}^{C\times H\times W}$. Afterwards, we calculate a semantic affinity matrix $\bm{W}_{i,j} = \frac{f_i \cdot f_j}{\lVert f_i \rVert_2 \lVert f_j \rVert_2}$. This affinity matrix represents a fully-connected undirected graph, i.e., an affinity graph, with cosine similarities as edge weights, on which we apply Normalized Cuts (NCut)~\cite{shi2000normalized} by solving $(Z-W)x = \lambda Zx$, where $Z$ is a $K \times K$ diagonal matrix with $z(i)=\sum_j W_{i,j}$. The goal of NCut is to find the eigenvector $x$ that matches the second smallest eigenvalue $\lambda$. The algorithm yields an eigenvalue decomposition of the graph, i.e., we obtain an eigenvalue for each node. This defines $\lambda_{\max} = \max_e |\lambda_e|$ as the semantically most forward and $\lambda_{\min} = \min_e |\lambda_e|$ as the semantically most backward point. 
MaskCut~\cite{wang2023cut} binarizes the graph with the threshold $\tau^\text{ncut}$ and cuts a bipartition $B$ from the graph, a process which is repeated for $N$ segmentation iterations per image. Our approach leverages this semantic bipartition as an initial semantic mask.  After each cut, we remove the nodes corresponding to the previously obtained segmentation from the affinity graph.  We find that in some cases, all objects in the scene are assigned an instance after $n<N$ iterations. This is indicated by the algorithm predicting the inverse of all previous segmentation masks combined, i.e., it predicts the "rest". If this occurs, we stop our pseudo-mask extraction for this image.

\begin{figure*}[!ht]
  \centering
    \includegraphics[width=\textwidth, page=3]{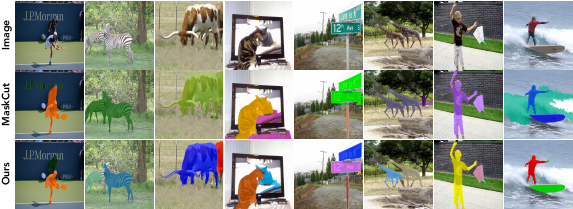}
    \caption{\textbf{Visualization of CutS3D Pseudo-Masks.} We showcase the capability of our pseudo-mask extraction pipeline. Our method is able to separate instances in 3D space, enabling the separation of same-class instances such as the humans playing tennis on the left. MaskCut~\cite{wang2023cut} only takes into account 2D, therefore it fails to separate the humans positioned behind each other.} 
    \label{fig:qualitative}
    \vspace{-4mm}
\end{figure*}
\subsection{LocalCut: Cutting Instances in 3D}
\label{sec:localcut}
To identify the most relevant semantic group of the scene, NCut relies on global semantic information of the scene, represented by the feature-based affinity graph. We argue that, to separate individual instances from a semantic group, local information is most effective. While the naive 2D connected component filter in MaskCut~\cite{wang2023cut} takes into account local properties, it can fail to identify the actual instance boundaries if the instances are connected in 2D and semantically similar.
We overcome this issue by leveraging 3D information in the form of a point cloud to segment instances along their actual boundaries in 3D space, which we illustrate in Figure~\ref{fig:pseudo-mask}. To achieve this, we first extract a depth map $D \in [0,1]$ from the input image using an off-the-shelf zero-shot monocular depth estimator, in this case ZoeDepth~\cite{bhat2023zoedepth}. We adapt this depth map to the resolution of the patch-level feature map $f$, and unproject it orthographically into a point cloud $P$, consisting of a set of points $\{ p_{1}, p_{2}, \ldots, p_{m} \}$. 
We leverage the previously described initial semantic bipartition $B$ and set the z-coordinates of points outside the semantic cut to a background level. To capture the local geometric properties of the 3D space, we construct a $k$-NN graph $G^{3D}=(V,E)$ on this pre-filtered point cloud with the edge  $e$ being assigned a weight $c$, the Euclidean distance between two points. 
This graph is effective for capturing local geometric structures within the semantic region, since a given point is only connected to its $k$ closest neighbors in 3D space.
We then threshold the graph with $\tau_\text{knn}$ and cut the instance from the semantic mask in 3D using MinCut on the point cloud~\cite{golovinskiy2009min}. MinCut aims to partition the graph into two disjoint subsets by minimizing the overall edge weights that need to be removed from the graph. More specifically, we use an implementation of Dinic's algorithm~\cite{dinic1970algorithm} to solve the maximum-flow problem, whose objective to maximize the total flow from the source node to the sink node is equal to the MinCut objective. 
For the algorithm to produce the desired output, source $s$ shall be set to the most foreground point and sink $t$ to the most background point. We exploit these two parameters to effectively connect the semantic space to the 3D space: Instead of defining foreground and background by analyzing the point cloud, we leverage information obtained through NCut on the semantic affinity graph and set $s = p_{\lambda_{\max}}$, i.e. the point at the maximum absolute eigenvalue, and $t = p_{\lambda_{\min}}$, i.e. the point at the minimum absolute eigenvalue. By definition, these are the points selected by NCut from the semantic space to be foreground and background. Like MaskCut~\cite{wang2023cut}, this mask is then further refined with a Conditional Random Field (CRF)~\cite{krahenbuhl2011crf}. We display a selection of generated pseudo-masks in Figure~\ref{fig:qualitative}.

\begin{figure*}[!th]
  \centering
    \includegraphics[width=\textwidth, page=2]{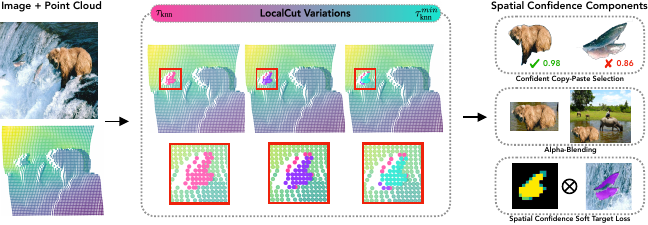}
    \caption{\textbf{Spatial Confidence Process.} We introduce Spatial Confidence maps to capture the quality of the 3D-semantic pseudo-mask extraction process. For this, we compute multiple cuts on the point clouds by varying $\tau_\text{knn}$, then accumulate and average the different masks. We use our Spatial Confidence for selecting confident masks and applying alpha-blending for copy-paste augmentation, as well as introducing a novel Spatial Confidence Soft Target Loss.} 
    \label{fig:spatial-confidence}
    \vspace{-4mm}
\end{figure*}

\subsection{Spatial Importance Sharpening}
\label{sec:spatial-importance}
As previously described, we partition the point cloud with a semantic mask as the basis for our LocalCut. To obtain this semantic mask in the first place, NCut~\cite{shi2000normalized} aims to find the most relevant region on the fully-connected semantic graph, so that the similarity within this region is maximized and the similarity between different regions is minimized. Since we can exploit 3D information of the scene, we propose to enrich the semantic graph and make it 3D-aware. We observe that the semantic cut can generate masks that do not fully capture the instance boundary, missing parts of the instance that are important for accurate segmentation. With the intuition that our LocalCut will profit from an improved semantic mask, we aim to sharpen the semantic similarities along the object border since we want to include the spatially important areas of the object in the semantic mask for LocalCut to find the accurate 3D boundary. If this region is not part of the mask, LocalCut will not be able to cut at the 3D boundary. 
To achieve this, we first compute a Spatial Importance map based on the available depth. The goal is to assign high Spatial Importance values to regions with high-frequency components, since those areas contain important information about where the actual object boundaries might be located. We take inspiration from work by Luft et al.~\cite{luft2006image} who introduce Spatial Importance as part of an unsharp masking technique for image enhancement. A Gaussian low-pass filter $G_{\sigma}$ is used to smooth out the high-frequency components. We apply Gaussian blurring to the depth map and subtract the original depth map to obtain the Spatial Importance $\Delta D$:
\begin{equation}
    \Delta D = \left| G_{\sigma} \ast D - D \right|
\end{equation}
where $G_\sigma \ast D$ denotes convolution with the Gaussian kernel. Further, we normalize $\Delta D$ to be in $\in [\beta, 1.0]$ :
\begin{equation}
    \Delta D_n = \frac{(1.0 - \beta) \cdot (\Delta D - \min \Delta D)}{ (\max \Delta D - \min \Delta D)} + \beta
\end{equation}
To now infuse our semantic affinity matrix with 3D, we re-sharpen the individual cosine similarities using with element-wise exponentiation
\begin{equation}
    \bm{W}_{i,j} = W_{i,j}^{\, {1-\Delta D_n}_{i,j}}, \quad \text{for } i, j = 1, \dots, N.
\end{equation}
Subtracting the original depth from the blurred depth reveals areas with rapid depth changes which have a high likelihood of representing a 3D object boundary. We set the lower bound $\beta=0.45$ based on empirical findings.
Our aim for this is to  sharpen importance of semantic similarities at object boundaries. 
In our ablation in Table~\ref{tab:abl-individual}, we show that this step to sharpen similarities in areas of high Spatial Importance in combination with LocalCut, which now can cut at the actual 3D boundary, leads to a significant boost in performance. We ablate $\beta$ in the supplementary.


\subsection{Spatial Confidence}
\label{sec:spatial-confidence}
While the generated pseudo-masks provide a useful learning signal, we make a similar observation as Wang et al.~\cite{wang2023cut}, that these masks profit from further processing to create a 'cleaner' signal. We also adopt their proposed solution of training a CAD on these pseudo-masks. To improve on this process, we further propose to extract information about the quality of the pseudo-masks using our 3D information. Specifically, we compute Spatial Confidence maps, which aim to capture the certainty of the individual patches within the final 3D cut. Figure \ref{fig:spatial-confidence} displays our process.

A central LocalCut parameter is the threshold $\tau_\text{knn}$. We observe that for objects with well-separated 3D boundaries, this parameter is insensitive and produces the same final mask at different values. In contrast, when the 3D boundary of the object is not well defined, the resulting segmentation mask will vary. We exploit this property to compute Spatial Confidence maps in an attempt to capture the quality of a given pseudo-mask, especially along its boundaries.
To compute the Spatial Confidence map $\text{SC}$ for a given instance, we linearly sample $T$ variations between $\tau_\text{knn}^{min}$ and $\tau_\text{knn}$ to perform LocalCut for each configuration. The resulting binary cuts $\text{BC}$ are accumulated and averaged to obtain the Spatial Confidence $\text{SC}$:
\begin{equation}
    \text{SC}_{i,j} = \frac{1}{T} \sum_{t=1}^{T} \text{BC}_{i,j}(t)
\end{equation}
We set the minimum confidence $SC_{ij}^{\min}$ in the map to $0.5$ and ablate this parameter in the supplementary. We generate a Spatial Confidence map for each generated pseudo-mask and utilize it in three different ways during CAD training.

\begin{table*}[!th]
    \centering
    \resizebox{\linewidth}{!}{
    \begin{tabular}{lcccccccccccc}
        \toprule
        & \multicolumn{6}{c}{COCO20K} & \multicolumn{6}{c}{COCOval2017} \\
        \cmidrule(lr){2-7} \cmidrule(lr){8-13}
        Method & AP$^{\text{box}}_{50}$ & AP$^{\text{box}}_{75}$ & AP$^{\text{box}}$ & AP$^{\text{mask}}_{50}$ & AP$^{\text{mask}}_{75}$ & AP$^{\text{mask}}$ & AP$^{\text{box}}_{50}$ & AP$^{\text{box}}_{75}$ & AP$^{\text{box}}$ & AP$^{\text{mask}}_{50}$ & AP$^{\text{mask}}_{75}$ & AP$^{\text{mask}}$ \\
        \midrule
        FreeSOLO~\cite{wang2022freesolo} & 9.7 & 3.2 & 4.1 & 9.7 & 3.4 & 4.3 & 9.6 & 3.1 & 4.2 & 9.4 & 3.3 & 4.3  \\
        UnSAM\textsuperscript{*}~\cite{wang2025unsam} & - & - & - & 6.2 & 3.1 & 3.3 & - & - & - & 6.6 & 3.3 & 3.6  \\
        ProMerge+~\cite{li2024promerge} & - & - & - & - & - & 9.0 & - & - & - & - & - & 8.9  \\
        CutLER~\cite{wang2023cut} & 22.4 & 11.9 & 12.5 & 19.6 & 9.2 & 10.0 & 21.9 & 11.8 & 12.3 & 18.9 & 9.2 & 9.7 \\
        CuVLER\textsuperscript{*}~\cite{arica2024cuvler} & 23.5 & 11.9 & 12.7 & 20.0 & 9.0 & 10.0 & 23.0 & 11.8 & 12.6 & 19.3 & 8.8 & 9.8  \\
        CutS3D (Ours) &\textbf{24.6} & \textbf{12.5} & \textbf{13.4} & \textbf{21.3} & \textbf{9.9} & \textbf{10.9} & \textbf{24.3} & \textbf{12.5} & \textbf{13.3} & \textbf{20.8} & \textbf{9.8} & \textbf{10.7} \\
        \midrule
        \textbf{\textcolor{darkgreen}{$\Delta$ vs. SOTA}} &\textbf{\textcolor{darkgreen}{+1.1}} & \textbf{\textcolor{darkgreen}{+0.6}} & \textbf{\textcolor{darkgreen}{+0.7}} & \textbf{\textcolor{darkgreen}{+1.3}} & \textbf{\textcolor{darkgreen}{+0.9}} & \textbf{\textcolor{darkgreen}{+0.9}} & \textbf{\textcolor{darkgreen}{+1.3}} & \textbf{\textcolor{darkgreen}{+0.7}} & \textbf{\textcolor{darkgreen}{+0.7}} & \textbf{\textcolor{darkgreen}{+1.5}} & \textbf{\textcolor{darkgreen}{+1.0}} & \textbf{\textcolor{darkgreen}{+0.9}} \\
        \bottomrule
    \end{tabular}
    }
    \caption{\textbf{Zero-Shot Unsupervised Instance Segmentation.} Our model is able to outperform the previous state-of-the-art (SOTA) in a zero-shot setting on COCO val2017 and COCO20K. CutS3D (Ours), CutLER, CuVLER, UnSAM and ProMerge+ are evaluated zero-shot.  \\ \textsuperscript{*} Results obtained using official checkpoint.}
    \label{tab:instance-seg}
    \vspace{-4mm}
\end{table*}

\noindent\textbf{Confident Copy-Paste Selection.}
Copy-paste augmentation~\cite{ghiasi2021simple} has been shown to be effective for training the CAD on the pseudo-masks~\cite{wang2023cut}. Therefore, we opt to also use this augmentation when training our model, but with a twist: Instead of randomly choosing which masks to copy, we select only the highest quality masks, as determined by our Spatial Confidence maps, to be augmented. For this, we average the confidence scores for the entire mask into a single score and use it to sort the masks within an image. This reduces the amount of ambiguous masks which are copied, leading to better performance as shown in Table~\ref{tab:abl-individual}.

\noindent\textbf{Confidence Alpha-Blending.}
We experiment with including alpha-blending augmentation when copy-pasting object masks. Standard copy-paste augmentation uses a binary mask to paste the selected object into a different image or location. Instead, we again make use of our Spatial Confidence mask to alpha-blend the uncertain regions of the object into the new image $I^{aug}$, making pixels $i,j$ partially transparent proportional to their confidence
\begin{equation}
    I^{\text{aug}}_{i,j} = \text{SC}_{i,j} \cdot I^{S}_{i,j} + (1 - \text{SC}_{i,j}) \cdot I^{T}_{i,j}
\end{equation}
with $I^S$ being the source image from which the object is copied and $I^T$ being the target image to-be-pasted-in. We set $\text{SC}_{i,j} = 0$ for regions other than the copied instance.
For areas with high confidence, the object is fully pasted, whereas the pixel values for regions with lower confidence are blended with those of the image that they are pasted into. Combined with the confidence-based mask selection, we observe this further improves performance.


\noindent\textbf{Spatial Confidence Soft Target Loss.}
To directly incorporate the pseudo-mask quality into the learning signal, we propose to modify the loss of our CAD. In CuVLER, Arica et al.~\cite{arica2024cuvler} propose a soft target loss, which essentially re-weights the loss for an entire mask by a scalar. Since our Spatial Confidence is computed for each region in the mask individually, we use it to introduce a Spatial Confidence Soft Target loss. Instead of multiplying the full mask loss by a scalar, we re-weight the loss for each mask region individually with its confidence score, performing a much more targeted operation to incorporate patch-level confidence.
We utilize a Cascade Mask R-CNN~\cite{cai2019cascade} as our CAD, which computes a binary cross-entropy (BCE) loss on the pseudo mask. For our Spatial Confidence Soft Target Loss, we re-weight each part of the loss using
\begin{equation}
    L_{\text{mask}} = \sum_{(i,j)} \text{SC}_{i,j} \cdot \text{BCE}(\hat{M}_{i,j}, M_{i,j})
\end{equation}
for each mask, with $\text{BCE}(\hat{M}_{i,j}, M_{i,j})$ being the binary cross entropy, $\hat{M}$ and $M$ being the predicted and target pseudo-mask, and $\text{SC}_{i,j}$ being the Spatial Confidence at $i,j$. In this way, the confidence of LocalCut in its pseudo-masks is more precisely reflected in the learning signal for the CAD. It is important to note that all loss values outside the confidence map are left unchanged, i.e. we set $\text{SC}_{i,j} = 1$.


 \section{Experiments}
\label{sec:formatting}

\noindent\textbf{Experiment Setup.} 
Following Wang et al.~\cite{wang2023cut}, we extract pseudo masks on the entire training split of ImageNet~\cite{russakovsky2015imagenet} (IN1K), consisting of roughly 1.3 million natural images. Each image is resized into $480 \times 480$ pixels and fed into the feature encoder $\mathcal{F}$. We utilize a DiffNCuts ViT-S/8~\cite{liu2024diffncuts} encoder, which is a DINO finetuned with differentiable NCut. Our approach also uses NCut for the initial semantic cut, making the network a natural choice.
Following, we adopt the choice of training a Cascade Mask R-CNN~\cite{cai2019cascade} with Spatial Confidence and DropLoss~\cite{wang2023cut} on the initial IN1K CutS3D pseudo masks. We keep the settings from CutLER largely the same and report hyperparameter configurations in the supplementary. After training on the pseudo-masks, we perform three rounds of self-training like CutLER~\cite{wang2023cut}.  We evaluate this model in our zero-shot evaluation settings on a variety of natural image datasets.
Additionally, we experiment with further training in-domain, as first proposed by Arica et al.~\cite{arica2024cuvler}. For this, we perform another round of self-training, but on COCO train2017~\cite{lin2014coco}. For a fair evaluation, we only compare this model against others which have conducted further self-training on the target domain.

\noindent\textbf{Datasets.} 
We evaluate our approach on an extensive suite of benchmarks and focus on natural image datasets, namely COCO val2017~\cite{lin2014coco}, COCO20K~\cite{lin2014coco} and LVIS~\cite{gupta2019lvis}, all containing instance and bounding box annotations. Additionally, we evaluate on datasets with only bounding box labels, namely Pascal VOC~\cite{everingham2010pascal}, Objects365~\cite{shao2019objects365} and KITTI~\cite{geiger2013kitti}.  


\begin{table*}[!th]
    \centering
    \resizebox{\linewidth}{!}{
    \begin{tabular}{l|cc|cc|cc|cc|cc|cc|cc}
        \toprule
        Method & \multicolumn{2}{c|}{Average} & \multicolumn{2}{c|}{COCOval2017} & \multicolumn{2}{c|}{COCO20K} & \multicolumn{2}{c|}{VOC} & \multicolumn{2}{c|}{Objects365} & \multicolumn{2}{c|}{KITTI} & \multicolumn{2}{c}{LVIS}  \\
        & AP$^{\text{box}}_{50}$ &  AP$^{\text{box}}$ & AP$^{\text{box}}_{50}$ &  AP$^{\text{box}}$ & AP$^{\text{box}}_{50}$ &  AP$^{\text{box}}$ & AP$^{\text{box}}_{50}$ &  AP$^{\text{box}}$ &
        AP$^{\text{box}}_{50}$ &  AP$^{\text{box}}$ & AP$^{\text{box}}_{50}$ &  AP$^{\text{box}}$ & AP$^{\text{box}}_{50}$ &  AP$^{\text{box}}$ \\
        \midrule
        CuVLER~\cite{arica2024cuvler}  & 21.3 & 11.3 & 23.0 & 12.6 & 23.5 & 12.7 & 39.4 & \textbf{22.3} & 21.6 & 10.9 & 11.8 & 4.6 & 8.5 & 4.5 \\
        CutLER~\cite{wang2023cut}  & 21.6 & 11.6 & 21.9 & 12.5 & 22.4 & 12.5 & 36.9 & 20.2 & 21.6 & 11.4 & 18.4 & 8.5 & 8.4 & 4.5  \\
        CutS3D (Ours) & \textbf{23.9} & \textbf{12.5} & \textbf{24.3} & \textbf{13.3} & \textbf{24.7} & \textbf{13.4} & \textbf{40.6} & 21.4 & \textbf{23.6} & \textbf{12.4} & \textbf{21.1} & \textbf{9.7} & \textbf{8.8} & \textbf{4.8} \\
        \midrule
        \textbf{\textcolor{darkgreen}{$\Delta$ vs. SOTA}} & \textbf{\textcolor{darkgreen}{+2.3}} & \textbf{\textcolor{darkgreen}{+0.9}} & \textbf{\textcolor{darkgreen}{+1.3}} & \textbf{\textcolor{darkgreen}{+0.7}} & \textbf{\textcolor{darkgreen}{+1.2}} & \textbf{\textcolor{darkgreen}{+0.7}} & \textbf{\textcolor{darkgreen}{+1.2}} & -0.9 & \textbf{\textcolor{darkgreen}{+2.0}} & \textbf{\textcolor{darkgreen}{+1.0}} & \textbf{\textcolor{darkgreen}{+2.7}} & \textbf{\textcolor{darkgreen}{+1.2}} & \textbf{\textcolor{darkgreen}{+0.3}} & \textbf{\textcolor{darkgreen}{+0.3}} \\
        \bottomrule
    \end{tabular}
    }
    \caption{\textbf{Zero-Shot Unsupervised Object Detection.} Our approach CutS3D outperforms competitors on multiple benchmarks, despite CutLER being further self-trained and CuVLER using a 6-model ensemble for pseudo-masks generation.}
    \label{tab:object-det}
    \vspace{-2mm}
\end{table*}

\begin{table}[!th]
  \vspace{-1mm}
  \centering
  \resizebox{\linewidth}{!}{
  \begin{tabular}{lcccccc}
    \toprule
    Dataset & \multicolumn{2}{c}{COCOval2017} & \multicolumn{2}{c}{COCO20K} & \multicolumn{2}{c}{LVIS} \\
    \cmidrule(lr){2-3} \cmidrule(lr){4-5} \cmidrule(lr){6-7}
    Method & AP$^{\text{mask}}_{50}$ & AP$^{\text{mask}}$ & AP$^{\text{mask}}_{50}$ & AP$^{\text{mask}}$ & AP$^{\text{mask}}_{50}$ & AP$^{\text{mask}}$ \\
    \midrule
    CuVLER~\cite{arica2024cuvler} & 20.4 & 10.4 & 21.6 & 10.7 & 7.2 & 3.8 \\
    CutS3D (Ours) & \textbf{21.8} & \textbf{11.4} & \textbf{22.4} & \textbf{11.6} & \textbf{8.1} & \textbf{4.5} \\
    \midrule
    \textbf{\textcolor{darkgreen}{$\Delta$ vs. SOTA}} & \textbf{\textcolor{darkgreen}{+1.4}} & \textbf{\textcolor{darkgreen}{+1.0}} & \textbf{\textcolor{darkgreen}{+0.8}} & \textbf{\textcolor{darkgreen}{+0.9}} & \textbf{\textcolor{darkgreen}{+0.9}} & \textbf{\textcolor{darkgreen}{+0.7}} \\
    \bottomrule
  \end{tabular}
  }
  \caption{\textbf{In-Domain Unsupervised Instance Segmentation.} Our method is able to outperform CuVLER after both models have been further self-trained on the COCO target domain.}
  \label{tab:in-domain}
  \vspace{-2mm}
\end{table}

\subsection{Unsupervised Instance Segmentation}
We first evaluate our IN1K-trained models on unsupervised instance segmentation in a zero-shot setting. Table~\ref{tab:instance-seg} shows comparison of our model on COCO20K and COCO val2017~\cite{lin2014coco} to the best performing models of many recent approaches for zero-shot unsupervised instance segmentation. 
Our model improves upon the best competitor by \textbf{+0.9} AP$^{\text{mask}}$ and \textbf{+1.5} AP$^{\text{mask}}_{50}$ on COCO val2017, and by \textbf{+0.9} AP$^{\text{mask}}$ and \textbf{+1.3} AP$^{\text{mask}}_{50}$ COCO20K.  Further, our approach also outperforms the CutLER baseline across all metrics by significant margins.
\subsection{Unsupervised Object Detection}
We further evaluate zero-shot object detection on a range of object detection datasets. In Table~\ref{tab:object-det}, we report the performance of our model and show that it improves upon best performing model on average by \textbf{+2.3} AP$^{\text{box}}_{50}$ and \textbf{+0.9} AP$^{\text{box}}$ across all datasets.
Similar to zero-shot unsupervised instance segmentation, our method again outperforms the best baseline method, CutLER~\cite{wang2023cut}, on all benchmarks. We also mostly outperform CuVLER~\cite{arica2024cuvler}, despite them using an ensemble of 6 different DINO ViTs for pseudo-mask extraction, while our method leverages only one feature extractor and 3D. This also showcases the effectiveness of 3D information in comparison to additional feature extractors.
\subsection{In-Domain Self-Training}
\label{sec:in-domain}
The previous results were obtained by training solely on ImageNet and then evaluating on other data domains. We also adopt the target-domain training setting from Arica et al.~\cite{arica2024cuvler} and further self-train our model for one more round on the COCO dataset. To obtain a pseudo ground truth, we use our zero-shot model to generate masks for COCO train2017. Table~\ref{tab:in-domain} compares our model to CuVLER, which has also been further self-trained on COCO. We outperform their method across all three versions of COCO. Our method improves by \textbf{+1.0} AP$^{\text{mask}}$ on the val2017 split and by \textbf{+0.9} AP$^{\text{mask}}$ on COCO20K. Notably, our model better captures the challenging fine-grained visual concepts of the long-tail LVIS benchmark where we outperform CuVLER by \textbf{+0.7} AP$^{\text{mask}}$. This shows our zero-shot model can produce accurate masks also outside of the ImageNet domain, and is effective for domain-specific training.


\section{Ablations}
\label{sec:formatting}
We ablate our proposed technical components by training the CAD on the generated IN1K pseudo-masks with DiffNCuts features and evaluate in a zero-shot manner on COCO val2017. For Table~\ref{tab:abl-individual}, we conduct only a single round of self-training (instead of 3 for main tables), to more closely reflect the effect of our contributions, without overemphasizing the effect of self-training.


\begin{table}[!thb]
\vspace{-1mm}
  \centering
  \resizebox{\linewidth}{!}{
  \begin{tabular}{lrrrr}
    \toprule
    & \multicolumn{4}{c}{COCO val2017} \\
    \cmidrule(lr){2-5}
    Method & AP$^{\text{box}}_{50}$ & AP$^{\text{box}}$ & AP$^{\text{mask}}_{50}$ & AP$^{\text{mask}}$ \\
    \midrule
    CutLER\textsuperscript{\textdagger} & 22.1 & 12.3 & 18.7 & 9.4 \\
    \hspace{1.5mm} + LocalCut (1) & 22.9 & 12.5 & 18.9 & 9.5 \\
    \hspace{1.5mm} + Spatial Importance (2) & 23.3 & 12.6 & 19.2 & 9.8 \\
    \hspace{1.5mm} + Spatial Confidence (3, 4, 5) & \textbf{23.9} & \textbf{13.0} & \textbf{20.1} & \textbf{10.2} \\
    \hspace{1.5mm} \textcolor{gray}{+ 3 Rounds Self-Training in Total} & \textcolor{gray}{24.3} & \textcolor{gray}{13.3} & \textcolor{gray}{20.8} & \textcolor{gray}{10.7} \\
    \bottomrule
  \end{tabular}
  }
  \caption{\textbf{Effect of our contributions.} We compare our individual contributions after one round of self-training. As base, we also train CutLER on DiffNCuts masks for 1 round.\\
  \textsuperscript{\textdagger}Results reproduced using the authors' official implementation.}
  \label{tab:abl-individual}
  \vspace{-4mm}
\end{table}

\noindent\textbf{Effect of our Individual Contributions.}
We investigate the effect of our contributions across both proposed backbones. In Table~\ref{tab:abl-individual}, we sequentially add the presented technical contributions and evaluate them by training first on pseudo-masks and then one round of self-training. Our baseline, CutLER, is trained with DINO-based pseudo-masks. Since we use DiffNCuts~\cite{liu2024diffncuts} instead, we re-train CutLER by using DiffNCuts features for MaskCut pseudo-mask generation and conduct one round of self-training, using the official author implementation. As can be observed, each added technical component improves our model over this baseline. A strong effect can be seen from the combination of LocalCut (1)  and Spatial Importance (2), since they amplify each other: With the improved semantics achieved through Spatial Importance sharpening, LocalCut can more often identify the 3D object boundary, leading to better performance. Spatial Confidence further improves performance and we present an analysis in the paragraph below.


\noindent\textbf{Equal Self-Training Rounds \& Backbone.} 
A crucial factor to our method's effectiveness is self-training. We make similar finding as Wang et al.~\cite{wang2023cut} that the CAD can extract the signal well from the pseudo-masks and refines it further with self-training. We conduct three round of self-training after the pseudo-mask training, just like CutLER. In Table~\ref{tab:abl-rounds}, we report results for the CAD trained with DINO- and DiffNCuts-based pseudo-masks for each round of self-training and with equal backbones and compare against CutLER. We find CutLER does not further benefit from the DiffNCuts backbone and its performance plateaus, since it still lacks to ability to separate instances with 3D.


\noindent\textbf{Spatial Confidence Components Analysis.} 
We further investigate the effect of our Spatial Confidence contributions. Therefore, we add each of the contributions that use Spatial Confidence individually to our CAD. We train the model only on the initial pseudo-masks without further self-training to best surface the effect, since self-training has the potential to blur the performance differences. As shown in Table~\ref{tab:abl-spatial-confidence}, selecting only confident masks for copy-paste augmentation leads to the biggest boost in performance. Adding alpha-blending with Spatial Confidence further improves results, while applying our Spatial Confidence Soft Loss further nudges up the performance even more.

\begin{table}[tb]
    \centering
    \resizebox{\linewidth}{!}{
    \begin{tabular}{ccccccc}
    \toprule
    AP$^{\text{mask}}$ & \multicolumn{3}{c}{DINO\cite{caron2021emerging}} & \multicolumn{3}{c}{DiffNCuts\cite{liu2024diffncuts}} \\ 
    \cmidrule(lr){2-4} \cmidrule(lr){5-7}
    Round & CutLER & Ours & \textcolor{darkgreen}{$\Delta$} & CutLER & Ours & \textcolor{darkgreen}{$\Delta$} \\ 
    \midrule
    1 & 8.8 &  \textbf{10.0} & \textbf{\textcolor{darkgreen}{+1.2}} & 9.4 & \textbf{10.2} & \textbf{\textcolor{darkgreen}{+0.8}} \\ 
    2 & 9.5 & \textbf{10.3} & \textbf{\textcolor{darkgreen}{+0.8}} & 9.6 & \textbf{10.4} & \textbf{\textcolor{darkgreen}{+0.8}} \\ 
    3 &  9.7 & \textbf{10.4} & \textbf{\textcolor{darkgreen}{+0.7}} & 9.8 & \textbf{10.7} & \textbf{\textcolor{darkgreen}{+0.9}} \\ 
    \bottomrule
    \end{tabular}
    }
    \caption{\textbf{Self-Training- and Backbone-fair Comparison.} Zero-Shot results for Ours \& CutLER on COCO val2017 after training the CAD on the initial pseudo-masks from DINO \& DiffNCuts for equal rounds. We outperform CutLER each round.}
    \label{tab:abl-rounds}
    \vspace{-2mm}
\end{table}

\begin{table}[tb]
    \centering
    \resizebox{\linewidth}{!}{
    \begin{tabular}{ccc|ccc}
        \toprule
        Confident Copy-Paste & Alpha Blend & Spatial Confidence Loss  & AP$^\text{mask}$ \\ 
        \midrule
        \xmark & \xmark & \xmark  & 8.5 \\ 
        \cmark & \xmark & \xmark  & 8.8 \\ 
        \cmark & \cmark & \xmark  & 9.0 \\ 
        \cmark & \cmark & \cmark & \textbf{9.1} \\
        \bottomrule
        \end{tabular}
        }
    \caption{\textbf{Spatial Confidence Component Analysis.} Zero-Shot results on COCO val2017 after training the Cascade Mask R-CNN on the initial pseudo-masks without self-training}
    \label{tab:abl-spatial-confidence}
    \vspace{-4mm}
\end{table}

\noindent\textbf{Depth Sources.} 
Recently, progress in zero-shot monocular depth estimators (MDEs) has led to many different models that generalize well across many domains. While some earlier approaches mix different datasets with sensor information \cite{ranftl2020midas, bhat2023zoedepth, ranftl2020towards} to train their model, more recent approaches using self-supervision \cite{spencer2023kick, spencer2024kick++, cecille2024groco} and training on synthetic depth \cite{ke2024marigold, he2024lotus, bochkovskii2024depthpro} have also gained traction. We select one approach from each of these categories to evaluate with our method: ZoeDepth~\cite{bhat2023zoedepth} is a finetuned MiDaS~\cite{ranftl2020midas} on a mix of metric depth datasets, Marigold~\cite{ke2024marigold} is a repurposed diffusion model to learn depth from synthetic data only, and Kick, Back \& Relax~\cite{spencer2023kick} is using self-supervision to learn depth from SlowTV videos. Lastly, we use depth from MiDaS (Small)~\cite{ranftl2020midas}, which produces depth of lower quality versus the other models. None of these models use datasets with human annotations. Each model predicts the depth for the IN1K training set. We generate pseudo-masks, only varying the depth source for the experiment. 
Using the pseudo-masks, we train the CAD with all contributions on the pseudo-masks without further self-training and report the zero-shot results on COCO val2017 in Table~\ref{tab:abl-depth}.
We observe that all depth estimators are equally suited for our method. 
\begin{table}[!ht]
  \centering
  \begin{tabular}{lcc}
    \toprule
    Method  & AP$^\text{mask}_{50}$ & AP$^\text{mask}$ \\ 
    \midrule
    ZoeDepth~\cite{bhat2023zoedepth} & \textbf{18.0} & \textbf{9.1}  \\
    Kick Back \& Relax~\cite{spencer2023kick} & 17.8 & 9.1 \\
    Marigold~\cite{ke2024marigold} &  17.8 & 8.9  \\
    MiDaS (Small)~\cite{ranftl2020midas} &  17.5 & 8.7  \\
    \bottomrule
  \end{tabular}
  \caption{\textbf{Different Depth Estimators} for our method. The top three produce detailed depth \& MiDAS predicts lower quality.
  }
  \label{tab:abl-depth}
  \vspace{-2mm}
\end{table}
\begin{table}[!tb]
\small
    \centering
    \begin{subtable}[t]{0.48\linewidth}
        \centering
        \begin{tabular}{lccc}
            \toprule
            $\tau_\text{knn}$ & .1 & .115 & .13 \\
            \midrule
            AP$_{\text{mask}}$ & 8.3 & \textbf{8.5} & 8.4 \\
            \bottomrule
        \end{tabular}
        \caption{Different values for $\tau_\text{knn}$.}
        \label{subtab:abl-tauknn}
    \end{subtable}
    \hfill
    \begin{subtable}[t]{0.48\linewidth}
        \centering
        \begin{tabular}{lcc}
            \toprule
            $\alpha$ & Scalar & Spatial \\
            \midrule
            AP$_{\text{mask}}$ & 8.8 & \textbf{9.0} \\
            \bottomrule
        \end{tabular}
        \caption{Alpha Blending Variations.}
        \label{subtab:abl-alpha}
    \end{subtable}
    \vspace{-2mm}
    \caption{\textbf{Further ablations.} We explore several variations of parameters and design aspects in our contributions.}
    \vspace{-4mm}
\end{table}
Lower quality depth does not degrade performance too much, while precise depth works best. Hence, we assume our approach can profit from future improved MDEs. However, our approach is still robust to imperfect depth. We show depth visualizations in the appendix.

\noindent\textbf{Computational Complexity.} In relation to the original MaskCut~\cite{wang2023cut}, our 3D operations add $+4\%$ overhead, split into depth prediction ($+.004\%$), LocalCut ($+0.8\%$), Spatial Importance ($+.001\%$) and Spatial Confidence ($+3\%$). Our modifications for detector training add $<1\%$ overhead. The inference cost of our detector is the same as CutLER~\cite{wang2023cut}.

\noindent\textbf{Further Ablations.} We further investigate the aspects of our contribution design choices. Table~\ref{subtab:abl-tauknn} shows variations for $\tau_\text{knn}$ in LocalCut, the CAD is trained on the pseudo-masks without Spatial Confidence to isolate the parameters effect. In Table~\ref{subtab:abl-alpha}, we experiment using the average confidence as $\alpha$ instead of the Spatial Confidence maps. We do not use the Spatial Confidence loss to highlight the effect. 




 \vspace{-1mm}
\section{Limitations}
Even when 3D information is available, our approach can struggle to extract accurate masks, if adjacent instances with similar semantics lack discernible 3D boundaries. Further, while we observe that our model can improve previous baselines for detection of smaller objects, this still remains an issue in unsupervised instance segmentation. We show a selection of failure cases in the supplementary. Further, our method's application is limited when 3D information cannot be obtained, as it is the case for some medical data. 
\vspace{-1mm}
\section{Summary}
We have introduced CutS3D, a novel approach for leveraging 3D information for 2D unsupervised instance segmentation. With our LocalCut, we cut along the actual 3D boundaries of instances, while Spatial Importance Sharpening enables clearer semantic relations in areas of high-frequency depth changes. Further, we propose the concept of Spatial Confidence to select only high-quality masks for copy-paste augmentation, alpha-blend the pasted objects and introduce a novel Spatial Confidence Soft Target Loss. CutS3D is able to achieve improved performance across multiple natural image benchmarks, in zero-shot settings and with further in-domain self-training. While CutS3D also improves results for in-domain training in comparison to other models, we believe extracting pseudo-masks directly in-domain without relying on ImageNet is a fruitful future direction.

 \newpage

\section*{Acknowledgements}
\small
We acknowledge the EuroHPC Joint Undertaking for awarding this project access to the EuroHPC supercomputer LEONARDO, hosted by CINECA (Italy) and the LEONARDO consortium through an EuroHPC Development Access call.
{
    \small
    \bibliographystyle{ieeenat_fullname}
    \bibliography{main}
}

\clearpage
\setcounter{page}{1}
\maketitlesupplementary
\appendix

\section{CutS3D Qualitative Results}
\begin{figure}[!ht]
    \centering
    \includegraphics[width=\linewidth, page=5]{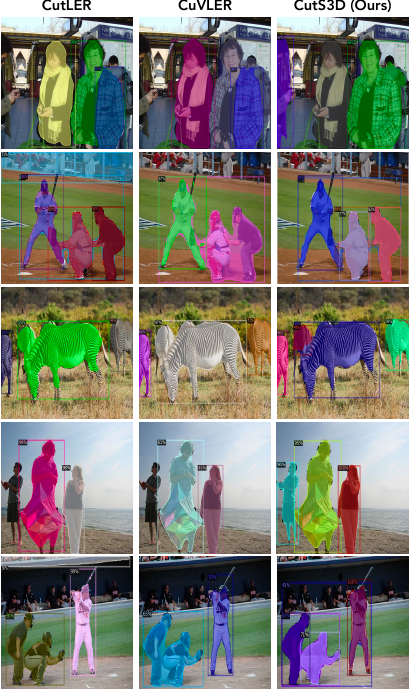}
    \caption{\textbf{More Qualitative Results.} We show COCO val2017 \cite{lin2014coco} predictions of our CutS3D zero-shot model and compare to zero-shot competitors, namely CutLER~\cite{wang2023cut} and CuVLER~\cite{arica2024cuvler}. Overall, we observe that the CutS3D Cascade Mask R-CNN~\cite{cai2019cascade} is able to better differentiate instances that are connected in 2D, e.g. located together in groups. On the other hand, the other two models often fail to separate such instances. }
    \label{fig:qualitative-teaser}
\end{figure}

\begin{figure*}[!h]
    \centering
    \includegraphics[width=\textwidth, page=1]{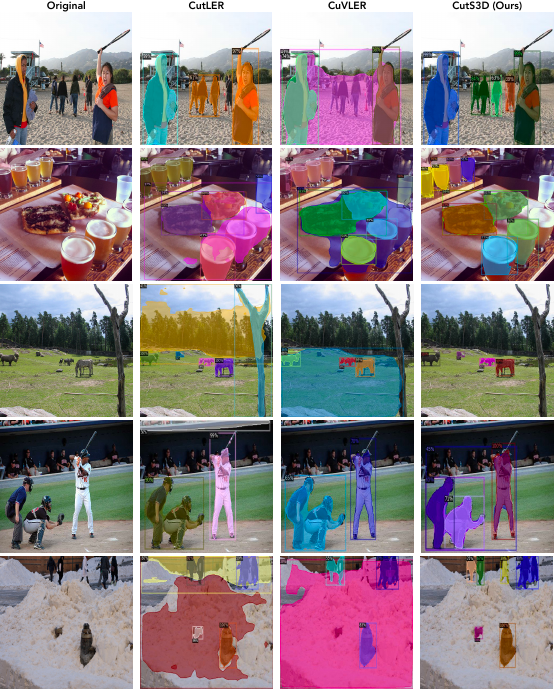}
    \caption{\textbf{More Qualitative Results.} We show further qualitative results on COCO val2017 from our zero-shot model and compare them to other zero-shot competitors for a fair comparison.}
    \vspace{-5mm}
    \label{fig:qualitative-suppl}
\end{figure*}

We show more qualitative examples of predictions from our CutS3D detector and compare it to other competitive methods in a zero-shot manner in Figure~\ref{fig:qualitative-teaser} and Figure~\ref{fig:qualitative-suppl}. Our approach shines for challenging examples with instances that are connected in 2D, such as the person holding the child or the baseball players standing together. The CutS3D CAD is also able to detect more instances, such as the additional zebra or the additional human at the bottom.

\section{Further Ablations}
\subsection{Spatial Importance Lower Bound}
We further ablate the effect for lower bound $\beta$ for our Spatial Importance maps  for the performance of our method. For this, we adopt the same evaluation protocol as in the main paper, i.e. we train our model only once on the generated ImageNet \cite{russakovsky2015imagenet} pseudo-masks. To isolate the effect of $\beta$, we train our model without Spatial Confidence. Table~\ref{subtab:abl-beta} reports the results on COCO val2017 \cite{lin2014coco} for $\beta$ variations.

\begin{table}[!h]
        \centering
        \begin{tabular}{lccc}
            \toprule
            $\beta$ & 0.3 & 0.45 & 0.6 \\
            \midrule
            AP$_{\text{mask}}$ & \textbf{8.5} & \textbf{8.5} & 8.3 \\
            \bottomrule
        \end{tabular}
        \vspace{-1.5mm}
        \caption{Different values for $\beta$.}
        \label{subtab:abl-beta}
\end{table}
\begin{table}[htb]
\small
    \begin{subtable}[t]{0.49\linewidth}
        \centering
        \begin{tabular}{lccc}
            \toprule
            $SC_{ij}^{\min}$ & 0.5 & 0.67 & 0.83 \\
            \midrule
            AP$_{\text{mask}}$ & \textbf{9.1} & 9.0 & 9.0 \\
            \bottomrule
        \end{tabular}
        \caption{$\text{SC}$ Lower Bound.}
        \label{subtab:abl-lowerbound}
    \end{subtable}%
    \hfill
    \begin{subtable}[t]{0.49\linewidth}
        \centering
        \begin{tabular}{lcc}
            \toprule
            & With & Without \\
            \midrule
            AP$_{\text{mask}}$ & 9.1 & 9.1 \\
            \bottomrule
        \end{tabular}
        \caption{$\text{SC}$ Maps Mask-Alignment.}
        \label{subtab:abl-maskalignment}
    \end{subtable}
    \vspace{-1.5mm}
    \caption{Further Spatial Confidence Components Ablations.}
    \label{tab:spatial-confidence-ablations}
\end{table}
\begin{table}[htb]
  \centering
  \begin{subtable}{0.48\linewidth}
  \begin{tabular}{lc}
    \toprule
    Contribution & AP$^\text{mask}$ \\ 
    \midrule
    \textcolor{gray}{CutLER} & \textcolor{gray}{9.4}  \\
    + SIS & 9.6  \\
    + LocalCut & 9.8  \\
    \bottomrule 
  \end{tabular}
  \caption{Spatial Importance Sharpening}
  \label{tab:sis}
  \end{subtable}%
  \hspace{2mm}
  \begin{subtable}{0.48\linewidth}
  \begin{tabular}{lc}
    \toprule
    Confidence  \hspace{4mm} & AP$^\text{mask}$ \\ 
    \midrule
    \textcolor{gray}{No Conf.} & \textcolor{gray}{9.8}  \\
    CRF & 10.0  \\
    Depth & 10.2  \\
    \bottomrule
  \end{tabular}
  \caption{Depth vs CRF Confidence}
  \label{tab:conf}
  \end{subtable}%
  \vspace{-1.5mm}
  \caption{Applying Spatial Importance Sharpening without LocalCut \& CRF scores as confidences.
  }
  \vspace{-2mm}
\end{table}

\begin{table*}[!ht]
\centering
\resizebox{\linewidth}{!}{
\begin{tabular}{clccccccccc|ccccccccc}
& Datasets & AP$^{\text{box}}_{50}$ & AP$^{\text{box}}_{75}$ & AP$^{\text{box}}$ & AP$^{\text{box}}_{\text{S}}$ & AP$^{\text{box}}_{\text{M}}$ & AP$^{\text{box}}_{\text{L}}$ & AR$^{\text{box}}_{1}$ & AR$^{\text{box}}_{10}$ & AR$^{\text{box}}_{100}$ & AP$^{\text{mask}}_{50}$ & AP$^{\text{mask}}_{75}$ & AP$^{\text{mask}}$ & AP$^{\text{mask}}_{\text{S}}$ & AP$^{\text{mask}}_{\text{M}}$ & AP$^{\text{mask}}_{\text{L}}$ & AR$^{\text{mask}}_{1}$ & AR$^{\text{mask}}_{10}$ & AR$^{\text{mask}}_{100}$ \\ 
\midrule
\multirow{6}{*}{\rotatebox[origin=c]{90}{Zero-Shot}}& COCO & 24.3 & 12.5 & 13.3 & 3.7 & 13.9 & 31.0 & 6.9 & 20.8 & 30.2 & 20.8 & 9.8 & 10.7 & 2.3 & 9.5 & 27.0 & 6.1 & 17.8 & 24.2 \\ 
& COCO20K & 24.6 & 12.6 & 13.4 & 4.0 & 14.0 & 30.7 & 6.9 & 20.9 & 30.4 & 21.3 & 9.9 & 10.9 & 2.4 & 9.9 & 26.7 & 6.1 & 17.8 & 26.7 \\
& LVIS & 8.8 & 4.2 & 4.8 & 2.3 & 8.8 & 15.7 & 2.4 & 9.9 & 17.9 & 7.1 & 3.6 & 3.9 & 1.4 & 5.7 & 13.8 & 2.2 & 8.5 & 14.6 \\ 
& VOC & 40.8 & 19.8 & 21.4 & 1.2 & 7.6 & 33.7 & 16.5 & 33.9 & 41.7 & - & - & - & - & - & - & - & - & - \\
& Objects365 & 23.6 & 11.2 & 12.4 & 2.7 & 11.0 & 22.5 & 3.2 & 16.5 & 30.8 & - & - & - & - & - & - & - & - & - \\
& KITTI & 21.1 & 7.6 & 9.7 & 0.1 & 5.7 & 22.9 & 7.0 & 17.5 & 25.3 & - & - & - & - & - & - & - & - & - \\
\midrule
\multirow{3}{*}{\rotatebox[origin=c]{90}{\scriptsize In-Domain}} & COCO & 24.7 & 12.5 & 13.3 & 3.6 & 13.4 & 29.8 & 6.7 & 21.1 & 31.5 & 21.8 & 10.4 & 11.4 & 2.5 & 10.5 & 27.5 & 6.1 & 19.0 & 27.0 \\
& COCO20K & 25.1 & 12.5 & 13.3 & 3.9 & 13.6 & 29.6 & 6.7 & 21.3 & 31.7 & 22.4 & 10.7 & 11.6 & 2.9 & 10.8 & 27.3 & 6.2 & 19.1 & 27.2 \\
    & LVIS & 9.6 & 4.6 & 5.2 & 2.5 & 8.9 & 15.3 & 2.4 & 10.4 & 29.5 & 8.1 & 4.1 & 4.5 & 1.9 & 6.5 & 14.2 & 2.3 & 9.5 & 17.2 \\ 
\bottomrule
\end{tabular}
}
\caption{\textbf{Full Results.} We report performance metrics for "Zero-Shot" and "In-Domain" settings for all datasets.}
\label{tab:full-results}
\end{table*}

\begin{table}[!ht]
  \centering
  \begin{tabular}{lcc}
    \toprule
    Model & AP$^\text{mask}_{50}$ & AP$^\text{mask}$ \\ 
    \midrule
    ZoeDepth \cite{bhat2023zoedepth} & 20.81 & 10.70  \\
    Kick Back \& Relax \cite{spencer2023kick}  & 20.72 & 10.69 \\
    \bottomrule
  \end{tabular}
  \vspace{-1.5mm}
  \caption{\textbf{3-Round Self-Training with Different Depth Models.} Our model fully trained with depth from ZoeDepth or Kick Back \& Relax.
  }
  \label{tab:depth}
  \vspace{-2mm}
\end{table}

\subsection{Spatial Confidence Lower Bound}
In Table~\ref{subtab:abl-lowerbound}, we explore setting different values as lower bound for our spatial confidence map. As minimum, we set $0.5$ and consecutively add $1/6\approx0.17$ since we make cuts at $6$ different thresholds. We find that our approach yields the best results at $SC_{ij}^{\min}=0.5$.

\subsection{Spatial Confidence Mask Alignment}
In an additional experiment, we investigate the effect of aligning the spatial confidence maps to pixel precision with further refinement instead of patch resolution. Using the two different resolutions for our spatial confidence soft target loss results in two different learning principles: While the coarse spatial confidence map encodes boundary confidence within a region, the refined map specifies exact borders with confidence. The detector either discovers or adjusts mask borders based on confidence. In Table~\ref{subtab:abl-maskalignment}, we find both approaches result in equal performance, rendering further refinement an unnecessary computational.

\subsection{Effect of Spatial Importance Sharpening}
In our method, we apply spatial importance sharpening (SIS) on the fully-connected feature graph before anything is cut. The goal of SIS is to sharpen the feature similarities of this fully-connected graph in areas where the edges are in the depth map. The intuition behind this is to increase the discriminatory effect of the feature similarities where they are the most important. Only adding SIS improves performance, as shown in Table~\ref{tab:sis} with a model trained on masks with only SIS applied for the semantic cut. We see LocalCut benefits from SIS on the semantic cut. Adding LocalCut in 3D now adds $+0.2$ (Table~\ref{tab:sis}) instead $+0.1$ without SIS (as shown in the main paper ablation). 

\subsection{Confidence from CRF}
We experiment with using soft masks from an alternative sources. One such alternative is the CRF, which outputs per-pixel probabilities for the refined instance mask. Naturally, these can also be plugged into our confidence components. Table~\ref{tab:conf} compares confidence from the CRF output vs. Spatial Confidence from depth. We find that using the CRF only slightly improves over not using any confidence. The results underline the value of deriving Spatial Confidence from 3D. A standalone object has high confidence, whereas for a candidate mask in an object group, the optimal 3D cut is less certain since the spatial boundary in 3D could be less-clear defined.

\begin{table}[!ht]
\centering
\begin{tabular}{lccc}
\toprule
Method & AP$^{\text{box}}$ & AP$^{\text{box}}_{50}$ & AP$^{\text{box}}_{75}$ \\
\midrule
MaskCut\textsuperscript{\textdagger}~\cite{wang2023cut}  & 8.4 & 15.0 & 8.0  \\
+ Ours  & \textbf{11.0} & \textbf{21.4} & \textbf{10.0}  \\
\bottomrule
\end{tabular}
\caption{\textbf{Pseudo-Mask Evaluation on ImageNet.} We evaluate the generated pseudo-masks on the ImageNet validation split \cite{russakovsky2015imagenet} for our baseline, MaskCut, and with our pseudo-mask contributions added (+ Ours). \textsuperscript{\textdagger}Results reproduced using the authors' official implementation. Since they do not provide pseudo-mask evaluation code, we use our own implementation only for this.}
\label{tab:pseudo-mask-results}
\end{table}

\subsection{Extended Depth Sources Ablation}
We extend our depth sources ablation in the main paper. Initially, we experiment with employing a self-supervised depth estimator, i.e. Kick Back \& Relax~\cite{spencer2023kick} that is trained on videos without any depth data. To evaluate the full extend of this alternative, we now perform 3-round self-training on the model already shown in the main paper, and, in Table~\ref{tab:depth}, present that it also achieves SOTA performance on COCO val2017. The model's performance quasi-matches the model performance trained with depth maps from ZoeDepth.

\subsection{Pseudo Mask Evaluation}
To train our CutS3D models, we first extract pseudo-masks on the ImageNet \cite{russakovsky2015imagenet} training split. Since ImageNet is a dataset that is mainly used for classification tasks, it lacks precise annotations for instance masks. Nevertheless, it comes with bounding box annotations, but those are constrained to one box per image. 
In an attempt to capture the abilities of our approach to extract a useful instance signal on ImageNet, we evaluate the our pseudo-mask extraction pipeline on the ImageNet validation split and report unsupervised object detection results in Table~\ref{tab:pseudo-mask-results}. To produce the numbers for our baseline, we use the official author implementation for CutLER~\cite{wang2023cut}for their pseudo-mask process called MaskCut. Both approaches use DiffNCuts~\cite{liu2024diffncuts} for feature extraction. As can be observed, our method scores higher across several metrics. This pseudo-mask advantage is also reflected in our presented results in the paper, where our trained CAD outperforms the baseline, CutLER \cite{wang2023cut}, with fewer self-training iterations.

\section{Full Results}
In addition to our results presented in the main paper, we report all metrics for the evaluated datasets in Table~\ref{tab:full-results}. This also includes different instance-size specific and instance recall metrics, which paint a more extensive picture.

\section{Further Visualizations}
\subsection{Pseudo-Mask Failure Cases}
\begin{figure}[!ht]
    \centering
    \includegraphics[width=\linewidth, page=4]{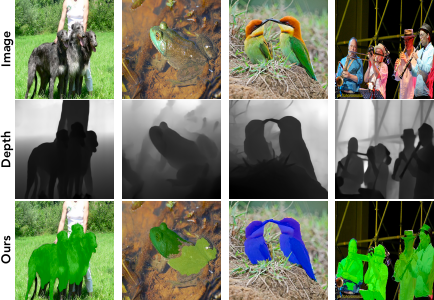}
    \caption{\textbf{Pseudo-Mask Failure Cases.} Our CutS3D pseudo-mask approach can struggle for objects with no discernable 3D boundary, such as the two birds sitting next to each other.}
    \label{fig:failure}
\end{figure}

While many of our generated pseudo-masks provide a reasonable segmentation of the instances in the scene, in some cases the predicted masks can be faulty or imprecise. Common cases are when objects are positioned next to each other with no discernible 3D boundary or simply when the initial semantic cut fails to find an instance. We therefore show examples of failure cases in Figure~\ref{fig:failure}.

\subsection{Depth Map Comparison}
\begin{figure*}[!h]
    \centering
    \includegraphics[width=\textwidth, page=2]{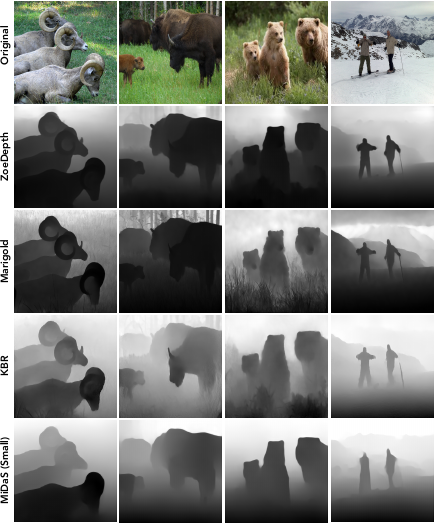}
    \caption{\textbf{Comparison of Different Monocular Depth Estimators.} Our visualizations qualitatively compare the depth maps predicted by ZoeDepth \cite{bhat2023zoedepth}, Marigold \cite{ke2024marigold}, Kick Back \& Relax \cite{spencer2023kick} and MiDaS \cite{ranftl2020midas}.}
    \label{fig:depth-comparison}
\end{figure*}

Our ablations in the main paper show that all evaluated zero-shot monocular depth estimators are suitable for our approach. Therefore, as part of Figure~\ref{fig:depth-comparison}, we show examples of predicted depth maps for all three models, namely ZoeDepth \cite{bhat2023zoedepth}, Marigold \cite{ke2024marigold}, and Kick Back \& Relax \cite{spencer2023kick}. 
Similar to the quantative evaluation, we observe that the depth maps from all three models are of similar high quality across a variety of scenes.

\subsection{Spatial Importance Maps}
\begin{figure*}[!h]
    \centering
    \includegraphics[width=\textwidth, page=3]{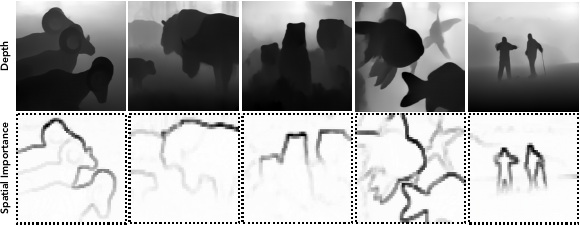}
    \caption{\textbf{Spatial Importance Examples.} We show Spatial Importance maps generated from depth maps predicted by ZoeDepth \cite{bhat2023zoedepth}.}
    \label{fig:spatialimportance-comparison}
\end{figure*}

As the contribution ablation reveals, sharpening the semantic affinity graph with Spatial Importance maps greatly improves the performance of our method. Therefore, we show further examples of Spatial Importance maps as part of Figure~\ref{fig:spatialimportance-comparison}. As can be observed, our Spatial Importance maps extract areas of high-frequency depth changes from the depth maps across various scenes.

\section{Method Details}
\subsection{Pseudo-Mask Extraction}
We detail our hyperparameters for pseudo-mask and Spatial Confidence map extraction in Table~\ref{tab:pseudo-details}. We perform $3$ iterations to identify instances. To extract Spatial Confidence, we linearly sample $6$ variations of $\tau_{\text{knn}}$. For our main results, we extract depth from ZoeDepth~\cite{bhat2023zoedepth} and features from DINO~\cite{caron2021emerging} or DiffNCuts \cite{liu2024diffncuts}.

\begin{table}[h]
  \centering
  \begin{tabular}{ll}
    \toprule
    Parameter & Value \\
    \midrule
    $N$ & $3$ \\
    $\tau_\text{NCut}$ & $0.13$ \\
    $\tau_\text{knn}$ & $0.115$ \\
    $\tau^{\min}_\text{knn}$ & $0.05$ \\
    $\beta$ & $0.45$ \\
    $T$ & $6$ \\
    Depth Model & ZoeDepth \cite{bhat2023zoedepth} \\
    \multirow{2}{*}{Backbones} & ViT-B/8 (DINO) \\
    & ViT-S/8 (DiffNCuts) \\
    \bottomrule
  \end{tabular}
  \caption{\textbf{Pseudo-Mask Extraction Hyperparameters.} We report the hyperparameters used for our LocalCut, Spatial Importance and Spatial Confidence processes.}
  \label{tab:pseudo-details}
\end{table}

\subsection{Initial Pseudo-Mask Training}
We report the hyperparameters used for the initial training of the Cascade Mask R-CNN on the pseudo-masks generated from ImageNet in Table~\ref{tab:pseudo-training-details}. For training the model, we largely follow the standard settings from CutLER \cite{wang2023cut} and train for 160K iterations. Due to additional memory needs from the Spatial Confidence maps, we reduce our batch size to $4$. For our ablations without spatial confidence, we increase it to $8$. Like CutLER \cite{wang2023cut}, we scale the copy-pasted masks between $0.3$ and $1.0$ to vary the resulting size of copied instances. We also initialize the model backbone from DINO~\cite{caron2021emerging} weights.
\begin{table}[h]
  \centering
  \begin{tabular}{ll}
    \toprule
    Component & Value \\
    \midrule
    Detector & Cascade Mask R-CNN \cite{cai2019cascade} \\
    Batch Size & $4$ \\
    Base Learning Rate & $1e^{-2}$ \\
    Optimizer & SGD \\
    Momentum & $0.9$ \\
    Weight Init. & DINO \cite{caron2021emerging} \\
    Warmup Iterations & 1K \\
    Total Iterations & 160K \\
    Copy-Paste Min. Ratio & $0.3$ \\
    Copy-Paste Max. Ratio & $1.0$ \\
    \bottomrule
  \end{tabular}
  \caption{\textbf{Initial Pseudo-Mask Training Cascade Mask R-CNN Hyperparameters.} We detail hyperparameters used for training the CAD on the generated pseudo-masks.}
  \label{tab:pseudo-training-details}
\end{table}

\subsection{Self-Training}
We further conduct self-training with the predicted masks from the initially trained Cascade Mask R-CNN and report our hyperparameters in Table~\ref{tab:self-training-details}. Since the CAD trained on the initial pseudo-masks cannot predict Spatial Confidence maps for self-training, we no longer have additional memory needs and hence increase the batch size to $8$. Further, we find the model converges after 80K iterations, partly due to its weights being initialized from the previously trained CAD. We further increase the minimum scale for copy-paste augmentation to $0.5$. Different from CutLER~\cite{wang2023cut}, we only conduct 1 round of self-training, saving computational costs.
For further in-domain self-training on COCO, we keep our settings largely the same and mainly reduce the total iterations to 14K since COCO is considerably smaller in size than ImageNet. We provide configuration files for all our trainings as part of the code.

\begin{table}[h]
  \centering
  \begin{tabular}{ll}
    \toprule
    Component & Value \\
    \midrule
    Detector & Cascade Mask R-CNN \cite{cai2019cascade} \\
    Batch Size & $8$ \\
    Base Learning Rate & $5e^{-3}$ \\
    Optimizer & SGD \\
    Momentum & $0.9$ \\
    Weight Init. & Previous Training \\
    Self-Training Rounds & 1 \\
    Warmup Iterations & 1K \\
    Total Iterations & 80K \\
    Copy-Paste Min. Ratio & $0.5$ \\
    Copy-Paste Max. Ratio & $1.0$ \\
    \bottomrule
  \end{tabular}
  \caption{\textbf{IN1K Self-Training Cascade Mask R-CNN Hyperparameters.} We report hyperparameters used for performing self-training of our CAD.}
  \label{tab:self-training-details}
\end{table}


\newpage

\end{document}